\pdfoutput=1

\documentclass[11pt]{article}

\usepackage{EMNLP2022}

\usepackage{times}
\usepackage{latexsym}
\usepackage{times}
\usepackage{latexsym}
\usepackage{microtype}
\usepackage{color}
\usepackage{bm}
\usepackage{amssymb}
\usepackage{amsfonts}
\usepackage{amsmath}
\usepackage{footmisc}
\usepackage{multirow}
\usepackage{multicol}
\usepackage{caption}
\usepackage{subcaption}
\usepackage{mflogo}
\usepackage{amsthm}
\usepackage{graphicx} 
\usepackage{booktabs}
\usepackage{xspace}
\usepackage{newfloat}
\usepackage{listings}
\usepackage[ruled,linesnumbered]{algorithm2e}
\usepackage[T1]{fontenc}

\usepackage[utf8]{inputenc}

\usepackage{microtype}
\newcommand{\paratitle}[1]{\vspace{1.5ex}\noindent\textbf{#1}}
\newcommand{\ie}{\emph{i.e.,}\xspace}

\newcommand{\eg}{\emph{e.g.,}\xspace}

\newcommand{\ignore}[1]{}
\usepackage[T1]{fontenc}

\usepackage[utf8]{inputenc}

\usepackage{microtype}

\usepackage{inconsolata}

\usepackage[noabbrev,capitalize]{cleveref}

\crefname{equation}{equation}{equations}   
\crefname{footnote}{footnote}{footnotes}   
\crefname{line}{line}{lines}               
\crefname{section}{\S}{\S\S}
\Crefname{section}{\S}{\S\S}    
\crefformat{section}{#2\S#1#3}  
\Crefformat{section}{#2\S#1#3}
\crefrangeformat{section}{\S\S#3#1#4--#5#2#6}
\Crefrangeformat{section}{\S\S#3#1#4--#5#2#6}

\title{SimANS: Simple Ambiguous Negatives Sampling for Dense Text Retrieval}

\author{
\setcounter{footnote}{1}
	Kun Zhou\textsuperscript{\rm{1},\rm{3}}\thanks{$^\dagger$ This work was done during internship at MSRA.},
	Yeyun Gong\textsuperscript{\rm{4}},
	Xiao Liu\textsuperscript{\rm{4}},
	\setcounter{footnote}{0}
	Wayne Xin Zhao\textsuperscript{\rm{2},\rm{3}}\thanks{$^*$ Corresponding author, email: batmanfly@gmail.com.},
	Yelong Shen\textsuperscript{\rm{5}}, 
	Anlei Dong\textsuperscript{\rm{5}}, \\
	\textbf{Jingwen Lu}\textsuperscript{\rm{5}}, 
	\textbf{Rangan Majumder}\textsuperscript{\rm{5}},
	\textbf{Ji-Rong Wen}\textsuperscript{\rm{2},\rm{3}},
	\textbf{Nan Duan}\textsuperscript{\rm{4}},
	\textbf{Weizhu Chen}\textsuperscript{\rm{5}} \\
	\textsuperscript{1}School of Information, Renmin University of China, \\
	\textsuperscript{2}Gaoling School of Artificial Intelligence, Renmin University of China, \\
	\textsuperscript{3}Beijing Key Laboratory of Big Data Management and Analysis Methods, \\
	\textsuperscript{4}Microsoft Research,
	\textsuperscript{5}Microsoft\\
}

\begin{document}
\maketitle
\begin{abstract}
Sampling proper negatives from a large document pool is vital to effectively train a dense retrieval model.
However, existing negative sampling strategies suffer from the uninformative or false negative problem.
In this work, we empirically show that according to the measured relevance scores, the negatives ranked around the positives are generally more informative and less likely to be false negatives.
Intuitively, these negatives are not too hard (\emph{may be false negatives}) or too easy (\emph{uninformative}). 
They are the ambiguous negatives and need more attention during training.
Thus, we propose a simple ambiguous negatives sampling method, SimANS, which incorporates a new sampling probability distribution to sample more ambiguous negatives.
Extensive experiments on four public and one industry datasets show the effectiveness of our approach.
We made the code and models publicly available in \url{https://github.com/microsoft/SimXNS}.
\end{abstract}

\section{Introduction}

Dense text retrieval, which uses low-dimensional vectors to represent queries and documents and measure their relevance, has become a popular topic~\cite{karpukhin2020dense,luan2021sparse} for both researchers and practitioners.
It can improve various downstream applications, \eg web search~\cite{brickley2019google,qiu2022dureader_retrieval} and question answer~\cite{izacard2021leveraging}. 
A key challenge for training a dense text retrieval model is how to select appropriate negatives from a large document pool (\ie negative sampling), as most existing methods use a contrastive loss~\cite{karpukhin2020dense,xiong2020approximate} to encourage the model to rank positive documents higher than negatives. 
However, the commonly-used negative sampling strategies, namely random negative sampling~\cite{luan2021sparse,karpukhin2020dense} (using random documents in the same batch) and top-$k$ hard negatives sampling~\cite{xiong2020approximate,DBLP:conf/sigir/ZhanM0G0M21} (using an auxiliary retriever to obtain the top-$k$ documents), have their limitations. 
Random negative sampling tends to select uninformative negatives that are rather easy to be distinguished from positives and fail to provide useful information~\cite{xiong2020approximate}, while top-$k$ hard negatives sampling may include false negatives~\cite{DBLP:conf/naacl/QuDLLRZDWW21}, degrading the model performance.

\ignore{
\begin{figure}[t]
  \centering
  \includegraphics[width=0.9\columnwidth]{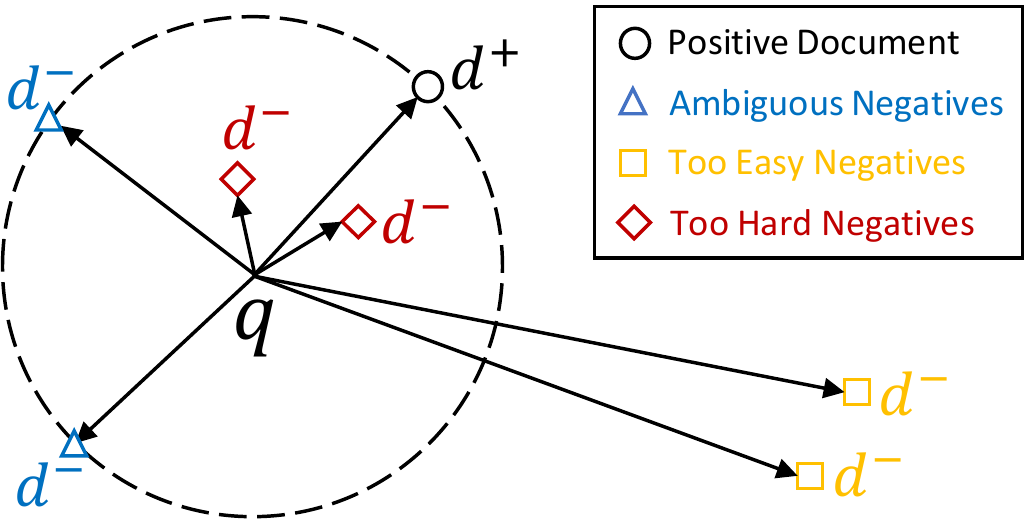}
  \caption{Examples of a positive example and its easy, hard and ambiguous negatives.}
  \label{fig:case}
\end{figure}
}


Motivated by these problems, we propose to sample the \emph{ambiguous negatives}~\footnote{We call them ambiguous negatives following the definition of ambiguous examples~\cite{swayamdipta2020dataset,meissner2021embracing}, referring to the instances that are neither too hard nor too easy to learn.} that are neither too easy (uninformative) nor too hard (potential false negatives).
Our approach is inspired by an empirical observation from experiments (in \cref{sec-motivation}) using gradients to assess the impact of data instances on deep models~\cite{koh2017understanding,pruthi2020estimating}: according to the measured relevance scores using the dense retrieval model, negatives that rank lower are mostly uninformative, as their gradient means are close to zero; negatives that rank higher are likely to be false negatives, as their gradient variances are significantly higher than expected.
Both types of negatives are detrimental to the convergence of deep matching models~\cite{xiong2020approximate,DBLP:conf/naacl/QuDLLRZDWW21}.
Interestingly, we find that the negatives ranked around positive examples tend to have relatively larger gradient means and smaller variances, indicating that they are informative and have a lower risk of being false negatives, thus probably being high-quality ambiguous negatives.


Based on these insights, we propose a \textbf{Sim}ple \textbf{A}mbiguous \textbf{N}egative \textbf{S}ampling method, namely \textbf{SimANS}, for improving deep text retrieval.
Our main idea is to design a sampling probability distribution that can assign higher probabilities to the ambiguous negatives while lower probabilities to the possible false and uninformative negatives, based on the differences of the relevance scores between positives and candidate negatives.
We also incorporate two hyper-parameters to better adjust the peak and density of the sampling probability distribution.
Our approach is simple and flexible, which can be easily applied to various dense retrieval models and combined with other effective techniques, \eg knowledge distillation~\cite{DBLP:conf/naacl/QuDLLRZDWW21} and adversarial training~\cite{zhang2021adversarial}.


To validate the effectiveness of SimANS, we conduct extensive experiments on four public datasets and one industrial dataset collected from Bing search logs.
Experimental results show that SimANS can improve the performance of competitive baselines, including state-of-the-art methods. 

\section{Preliminary}
\paratitle{Dense Text Retrieval.}
Given a query $q$, the dense text retrieval task aims to retrieve the most relevant top-$k$ documents $\{d_{i}\}^{k}_{i=1}$ from a large candidate pool $\mathcal{D}$.
To achieve it, the dual-encoder architecture is widely used due to its efficiency~\cite{reimers2019sentence,karpukhin2020dense}. It consists of a query encoder $E_{q}$ and a document encoder $E_{d}$ to map the query $q$ and document $d$ into $k$-dimensional dense vectors $\textbf{h}_{q}$ and $\textbf{h}_{d}$, respectively. Then, the semantic relevance score of $q$ and $d$ can be computed using dot product as
\begin{equation}
    s(q,d)=\textbf{h}_{q} \cdot \textbf{h}_{d}.
\end{equation}
Recent works mostly adopt pre-trained language models (PLMs)~\cite{devlin2019bert} as the two encoders, and utilize the representations of the \texttt{[CLS]} token as dense vectors.

\paratitle{Training with Negative Sampling.}
The training objective of dense text retrieval task is to pull the representations of the query $q$ and relevant documents $\mathcal{D}^{+}$ together (as positives), while pushing apart irrelevant ones $\mathcal{D}^{-}=\mathcal{D}\setminus \mathcal{D}^{+}$ (as negatives).
However, the irrelevant documents are from a large document pool, which would lead to millions of negatives.
To reduce the unreachable training cost, negative sampling has been widely used.
Previous works either randomly sample negatives~\cite{karpukhin2020dense}, or select the top-$k$ hard negatives ranked by BM25 or the dense retrieval model itself~\cite{xiong2020approximate,DBLP:conf/naacl/QuDLLRZDWW21}, denoted as $\widetilde{\mathcal{D}}^{-}$.
Then, the optimization objective can be formulated as:
\begin{equation}
\small
    \theta^{*}=\arg\min_{\theta} \sum_{q}\sum_{d^{+}\in\mathcal{D}^{+}}\sum_{d^{-}\in\widetilde{\mathcal{D}}^{-}} L(s(q,d^{+}),s(q,d^{-})), 
\end{equation}
where $L(\cdot)$ is the loss function.

\section{Motivation Study}
\label{sec-motivation}
We first analyze the uninformative and false negative problems from the perspective of gradients. 
Then, we perform an empirical study to test how gradients of negatives change \emph{w.r.t.} ranks according to measured relevance scores using a dense retrieval model, and find that the gradients of negatives ranked near positives have relatively larger means and smaller variances.

\subsection{Analysis for Gradients of Negatives}
\label{sec-analysis}

Existing dense retrieval methods~\cite{karpukhin2020dense,xiong2020approximate} commonly incorporate the binary cross entropy (BCE) loss to compute gradients~\footnote{In this work, we perform the analysis using BCE loss, and such analysis can also be extended to other loss functions.}, where the relevance scores of a positive and sampled negatives are usually normalized by the softmax function. In this way, the gradients of model parameters $\theta$ are computed by
\begin{equation}
\small
\nonumber
    \bigtriangledown_{\theta}l(q,d)=\begin{cases} (s_{n}(q,d)-1)\bigtriangledown_{\theta}s_{n}(q,d) \hspace{0.3cm} if~d\in \mathcal{D}^{+} \\ 
    s_{n}(q,d)\bigtriangledown_{\theta}s_{n}(q,d) \hspace{1.1cm} if~d\in \mathcal{D}^{-} \end{cases}
\end{equation}
where $s_{n}(q,d)$ is the normalized value of $s(q,d)$ and is within $[0,1]$.
Based on it, we review the gradients of uninformative and false negatives.
Uninformative negatives can be easily distinguished by dense retrieval models, and are more likely to be selected by random sampling~\cite{xiong2020approximate}.
As their normalized relevance scores are usually rather small, \ie $s_{n}(q,d)\longrightarrow 0$, their gradient means will be bounded into near-zero values, \ie $\bigtriangledown_{\theta}l(q,d)\longrightarrow 0$.
Such near-zero gradients are also uninformative and contribute little to model convergence.
False negatives are usually semantically similar to positives, and are more likely to be selected by top-$k$ hard negatives sampling~\cite{DBLP:conf/naacl/QuDLLRZDWW21}.
Therefore, for the gradients of false negatives and positives, the right terms $\bigtriangledown_{\theta}s_{n}(q,d)$ may be similar, while the left terms are greater than zero and less than 0, respectively.
As a result, the variance of gradients will be larger, which may cause the optimization of parameters to be unstable. 
Furthermore, existing works~\cite{katharopoulos2018not,johnson2018training} have theoretically proved that larger gradient variance is detrimental to model convergence.

\begin{figure}[t]
  \centering
  \includegraphics[width=1\columnwidth]{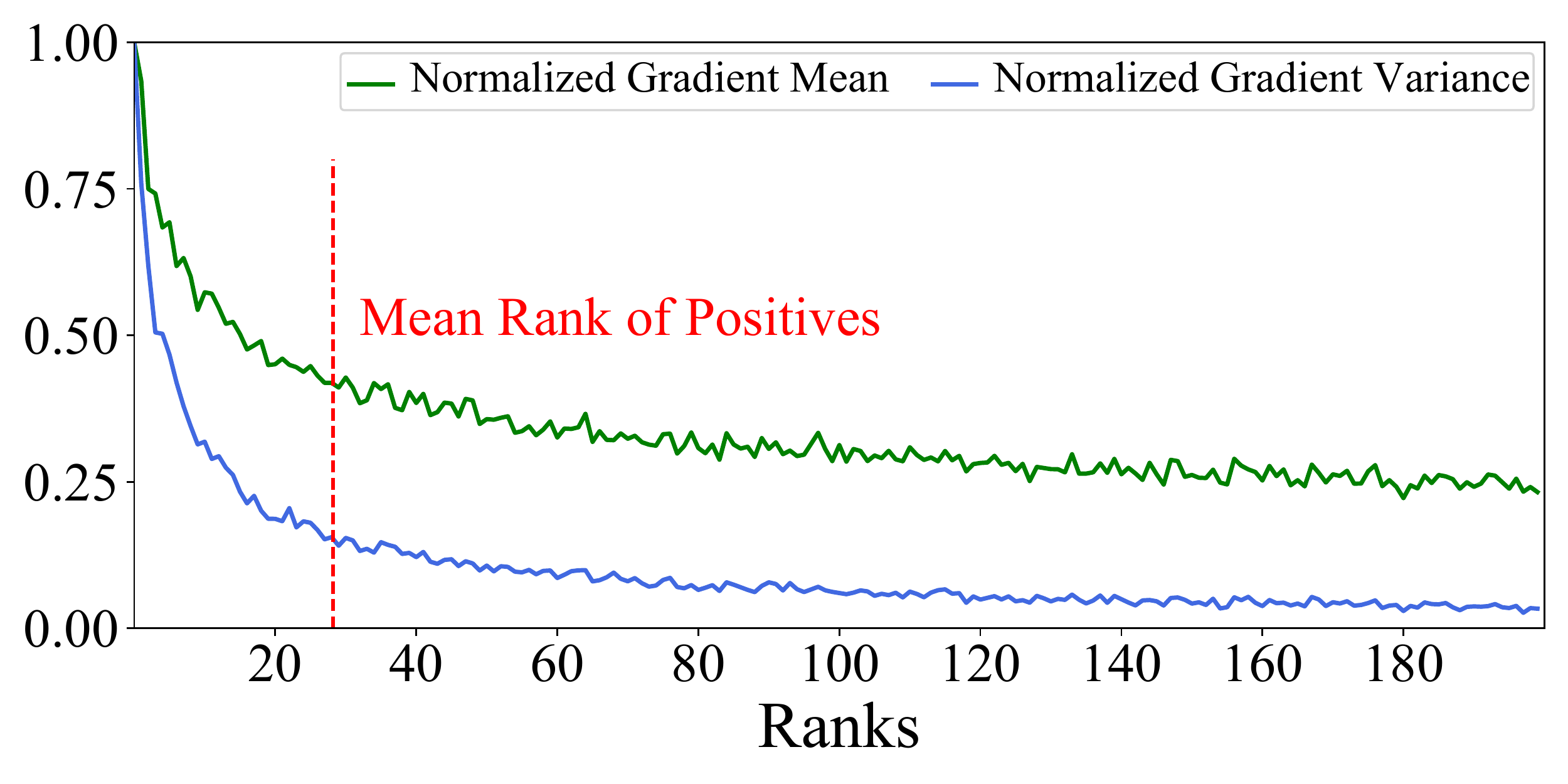}
  \caption{The mean and variance of gradients change curves \emph{w.r.t.} the ranks of negatives on MS-MARCO Passage Ranking dataset using AR2~\cite{zhang2021adversarial}.}
  \label{fig:motivation}
\end{figure}

\subsection{Empirical Study on Gradients of Negatives \emph{w.r.t.} Relevance Scores}
Although we have analyzed that the harmful influence of uninformative and false negatives derives from the smaller means and larger variances of gradients respectively, it is time-consuming to compute gradients of all candidate negatives to identify and remove them.
Here, we empirically study if the query-document relevance scores can be leveraged to avoid sampling these harmful negatives.

\paratitle{Experimental Setup.}
We use AR2~\cite{zhang2021adversarial} as the retrieval model and investigate its gradients on the development set of MS-MARCO Passage Ranking dataset~\cite{nguyen2016ms}.
Concretely, for each query, we rank all negatives according to their relevance scores, and compute the means and variances of gradients of all negatives in the same rank~\footnote{As AR2 adopts ERNIE-2.0~\cite{sun2020ernie} as the backbone that has millions of parameters, we only compute gradients on the parameters of its last layer for efficiency.}.
To better show the tendency \emph{w.r.t.} ranks of relevance scores, we normalize the means and variances of gradients by dividing the maximum values, and only report the results of top 200 ranked negatives.

\paratitle{Results and Findings.}
As shown in \cref{fig:motivation}, the mean and variance of gradients will gradually decrease with the increase of the rank. 
Despite that, the gradient means of the top 200 negatives are still in the same order of magnitude ($1.0\longrightarrow 0.25$), while the gradient variances of the top 10 ranked negatives are significantly larger than others.
The reason is that the higher-ranking negatives have larger probabilities to be false negatives.
Besides, a surprising finding is that the mean rank of positives is approximate the boundary point of the high gradient variance part and the negatives near it can produce relatively larger gradient means and lower gradient variances.
It means that they are high-quality \emph{ambiguous negatives} that can balance the informativeness and the risk of being false negatives. 
Therefore, it is promising to rely on the relevance scores of positives and candidate negatives to devise more effective negative sampling methods for training dense retrieval models.

\section{Approach}
Based on the findings in \cref{sec-motivation}, we conjecture that the ambiguous negatives ranked near positives according to relevance scores are high-quality negatives, as they are neither too easy (uninformative) nor too hard (may be false negatives).
Therefore, we propose a simple ambiguous negative sampling method, namely SimANS.

\subsection{Ambiguous Negative Sampling}
To focus on sampling ambiguous negatives, we design a new sampling probability distribution that can estimate the influence of each negative using the dense retrieval models. 
As follows, we first devise a general sampling distribution and then propose its simple and efficient implementation.

\paratitle{General Sampling Distribution.}
We draw the following conclusions from our results about how to choose a good sampling probability distribution for negatives:
(1) Negatives that are clearly irrelevant and have low relevance scores should be sampled less frequently;
(2) Negatives that are highly relevant and have high relevance scores should also be sampled less frequently, because they are more likely to be positives in disguise;
(3) Negatives that are uncertain and have relevance scores similar to positives should be sampled more frequently, because they provide useful information and have a lower chance of being false negatives.
We propose a general formula for negative sampling probability that reflects these principles:

\begin{equation}
\small
\label{eq-general}
p_{i} \propto f(|s(q,d_{i})-\bar{s}(q,d^{+})-b|), \forall~d_{i} \in \mathcal{D}\setminus \mathcal{D}^{+},
\end{equation}
where $f(\cdot)$ is a function to determine the tendency of the probability distribution, $b$ is a hyper-parameter to control the peak of the distribution, $\bar{s}(q,d^{+})$ is the mean relevance score of all positives with the query.
$f(\cdot)$ should be a monotone decreasing function (\eg $e^{-x}$).
In this way, the negatives with the relevance scores close to positives can be assigned with larger probabilities, while others with smaller or larger scores will be punished with smaller probabilities.
Such a distribution can satisfy the required three characteristics.

\paratitle{Simple Negative Sampling Distribution.}
We rely on several empirical priors to determine a simple and efficient implementation of the above sampling probability distribution.
Generally, the relevance scores of positives and negatives are bounded by the modulus of dense vectors, hence they are mostly in a same order of magnitude.
To ensure that the probabilities of ambiguous negatives should be significantly larger than other ones, we choose the exponential function to implement $f(\cdot)$.
As a large proportion of negatives from $\mathcal{D}\setminus \mathcal{D}^{+}$ are uninformative ones, their smaller relevance scores would lead to near-zero probabilities using the exponential function.
Therefore, we can reduce the computation cost by narrowing the negative candidates into the top-$k$ ranked negatives $\widetilde{\mathcal{D}}^{-}$.
In addition, to further reduce the cost, we also replace the mean relevance score of all positives $\bar{s}(q,d^{+})$ by the score of a randomly sampled positive $s(q,\tilde{d}^{+})$.
Finally, we can reformulate the sampling probability distribution in \cref{eq-general} as:
\begin{equation}
\small
\label{eq-simans}
p_{i} \propto \exp{(-a(s(q,d_{i})-s(q,\tilde{d}^{+})-b)^{2})}, 
\forall~d_{i} \in \widetilde{\mathcal{D}}^{-}, 
\end{equation}
where $a$ is a hyper-parameter to control the density of the distribution, $\tilde{d}^{+}\in \mathcal{D}^{+}$ is a randomly sampled positive,
$\widetilde{\mathcal{D}}^{-}$ is the top-$k$ ranked negatives.
In this way, the complexity of computing the sampling probability distribution will be reduced into $O(k)$, where $k \ll |\mathcal{D}|$ and we set it to 100.

\subsection{Overview and Discussion}

\begin{algorithm}[t]
\small
\caption{The algorithm of SimANS.}
\label{algorithm}
\LinesNumbered
\KwIn{Queries and their positive documents $\{(q,\mathcal{D}^{+})\}$, document pool $\mathcal{D}$, pre-learned dense retrieval model $M$}
Build the ANN index on $\mathcal{D}$ using $M$. \\
Retrieve the top-$k$ ranked negatives $\widetilde{\mathcal{D}}^{-}$ for each query with their relevance scores $\{s(q,d_{i})\}$ from $\mathcal{D}$. \\
Compute the relevance scores of each query and its positive documents $\{s(q,\mathcal{D}^{+})\}$. \\
Generate the sampling probabilities of retrieved top-$k$ negatives $\{p_{i}\}$ for each query using Eq.~\ref{eq-general}. \\
Construct new training data $\{(q,\mathcal{D}^{+},\widetilde{\mathcal{D}}^{-})\}$. \\
\While{$M$ has not converged}{
Sample a batch from $\{(q,\mathcal{D}^{+},\widetilde{\mathcal{D}}^{-})\}$. \\
Sample ambiguous negatives for each instance from the batch according to $\{p_{i}\}$. \\
Optimize parameters of $M$ using the batch and sampled negatives. \\
}
\end{algorithm}

\paratitle{Overview.}
Given a mini-batch, SimANS contains three major steps to obtain the ambiguous negatives.
The first step is the same as previous top-$k$ hard negatives sampling methods~\cite{xiong2020approximate,DBLP:conf/naacl/QuDLLRZDWW21} that select the top-$k$ ranked negatives $\widetilde{\mathcal{D}}^{-}$ from the candidate pool $\mathcal{D}\setminus\mathcal{D}^{+}$ using an ANN search tool (\eg FAISS~\cite{johnson2019billion}).
Second, we compute the sampling probabilities for all the top-$k$ negatives using \cref{eq-simans}. To reduce the time cost, we can pre-compute them in the first step.
Finally, we sample the ambiguous negatives \emph{w.r.t.} their sampling probabilities.
We present the overall algorithm in Algorithm~\ref{algorithm}.
Note that our proposed SimANS is a negative sampling method and applicable to a variety of dense retrieval methods.

\paratitle{Relationship with Other Methods.}
SimANS aims to sample the ambiguous negatives that rank close to the positives according to relevance scores for improving the training of dense retrieval models.
It is a general framework that several previous negative sampling methods can be included:

$\bullet$ \textbf{Choosing negative examples randomly} means picking them from a big collection of documents with equal chances for each one. We can also use our method to do this by setting $b=s(q,d_{i})-s(q,\tilde{d}^{+})$ and making $\widetilde{\mathcal{D}}^{-}$ include all the documents in the collection. But this is not a good idea, because most of the documents in the collection are not relevant to the query and do not help us learn from the feedback. They are easy to sample but not useful for training.

$\bullet$ \textbf{Top-$k$ hard negatives sampling} utilizes an auxiliary retriever (\eg BM25~\cite{karpukhin2020dense} or DPR~\cite{xiong2020approximate}) to rank all negative candidates and pick the top-$k$ ones as negatives. By setting $b=-s(q,\tilde{d}^{+})$ and $a=-\inf$, our method can also produce extremely large probabilities to the top-$k$ negatives.
Whereas, the top-$k$ ones have a higher risk to be false negatives, which are harmful to convergence.

\begin{table}[t]
\centering
\small
\begin{tabular}{l|cccc}
\toprule
\multicolumn{1}{l|}{\textbf{Datasets}} & \textbf{Training} & \textbf{Dev} & \textbf{Test} & \textbf{Documents} \\
\hline
NQ & 58,880 & 8,757 & 3,610 & 21,015,324 \\
TQ & 60,413 & 8,837 & 11,313 & 21,015,324 \\
MS Pas & 502,939 & 6,980 & - & 8,841,823 \\
MS Doc & 367,013 & 5,193 & - & 3,213,835 \\ 
Bing & 1,861,102 & 8,013 & - & 5,335,927 \\
\bottomrule
\end{tabular}
\caption{Statistics of the five text retrieval datasets. } 
\label{tab-dataset}
\end{table}

\begin{table*}[t]
\begin{center}
\resizebox{\textwidth}{!}{
\begin{tabular}{l|ccc|ccc|ccc}
\toprule
\multicolumn{1}{l|}{\multirow{2}{*}{\textbf{Method}}} & \multicolumn{3}{c|}{\textbf{NQ}} & \multicolumn{3}{c|}{\textbf{TQ}} & \multicolumn{3}{c}{\textbf{MS Pas}}\\ 
 & \textbf{R@5} & \textbf{R@20} & \multicolumn{1}{l|}{\textbf{R@100}}& \textbf{R@5} & \textbf{R@20} & \multicolumn{1}{l|}{\textbf{R@100}} & \textbf{MRR@10} & \textbf{R@50} & \textbf{R@1k} \\
 \hline
BM25~\citep{yang2017anserini} & - & 59.1 & 73.7 & - & 66.9 &  76.7 & 18.7 & 59.2 &  85.7 \\
GAR~\citep{DBLP:conf/acl/MaoHLSG0C20} & 60.9 & 74.4 & 85.3 & 73.1 &  80.4 & 85.7 & - & - &  - \\
doc2query~\citep{nogueira2019document} & - & - &  - & - & - &  - &  21.5 & 64.4 & 89.1 \\
DeepCT~\citep{dai2019deeper} & - & - &  - & - & - &  - & 24.3 & 69.0 &  91.0 \\
docTTTTTquery~\citep{nogueira2019doc2query} & - & - &  - & - & - &  - & 27.7 & 75.6 & 94.7 \\
\hline
DPR~\citep{karpukhin2020dense} & - &  78.4 & 85.3 & -  & 79.3 &  84.9 & - & - &  - \\
ANCE~\citep{xiong2020approximate} & 71.8 & 81.9 & 87.5 & -  & 80.3 & 85.3& 33.0  & 81.1 &  95.9 \\
COIL~\citep{DBLP:conf/naacl/GaoDC21}  & - & - &  - & - & - &  - & 35.5 & - & 96.3  \\
ME-BERT~\citep{luan2021sparse}  & - & - &  - & - & - &  -  & 33.8 &  - &  - \\
Joint top-$k$~\citep{sachan2021end} &  72.1  &  81.8 &  87.8 & 74.1 &  81.3  & 86.3 & - & - &  - \\
Individual top-$k$~\citep{sachan2021end} & 75.0  &   84.0 &  89.2 & 76.8 &  83.1  & 87.0 & - & - & - \\
RocketQA~\citep{DBLP:conf/naacl/QuDLLRZDWW21} & 74.0 &  82.7 &  88.5& - & -  & - &  37.0 & 85.5 & 97.9 \\
RDR~\citep{yang2020retriever} & - & 82.8 & 88.2 & -  & 82.5  & 87.3& - & - & - \\
RocketQAv2~\citep{ren2021rocketqav2} & 75.1 & 83.7 & 89.0 & & & & 38.8 & 86.2 & 98.1 \\
PAIR~\citep{DBLP:conf/acl/RenLQLZSWWW21} & 74.9 &  83.5 &  89.1 & - & -  & - & 37.9& 86.4 &98.2\\
DPR-PAQ~\citep{ouguz2021domain} & 74.2 &  84.0 & 89.2  & -  & -  & - & 31.1 & - & -  \\
Condenser~\citep{gao2021your}  & - &   83.2 &  88.4  & -  &  81.9 & 86.2 & 36.6 & - &  97.4 \\
coCondenser~\citep{gao2021unsupervised}   & 75.8 &   84.3 &  89.0  & 76.8  &  83.2  & 87.3 & 38.2 & - &  98.4 \\
ERNIE-Search~\citep{lu2022ernie} & 77.0 & 85.3 & 89.7 & - & - & - & \underline{40.1} & 87.7 & 98.2 \\
AR2~\citep{zhang2021adversarial} & \underline{77.9}	& \underline{86.0} 	& \underline{90.1}  & \underline{78.2} & \underline{84.4} & \underline{87.9} & 39.5 & \underline{87.8} &  \underline{98.6}\\
\hline
AR2+SimANS & \textbf{78.6} & \textbf{86.2} & \textbf{90.3} & \textbf{78.6} & \textbf{84.6} & \textbf{88.1} & \textbf{40.9} & \textbf{88.7} & \textbf{98.7} \\
\bottomrule
\end{tabular}
}
\caption{Performance on the test sets of NQ and TQ, and the development set of MS Pas.
The results of baselines are from original papers. 
The best and second-best methods are marked in bold and underlined, respectively.}
\label{tab.mainresult}
\end{center}
\end{table*}

\section{Experiments}
\subsection{Experimental Setting}

We extensively evaluate SimANS by conducting experiments on three public passage retrieval datasets: Natural Question~(NQ)~\cite{DBLP:journals/tacl/KwiatkowskiPRCP19}, Trivia QA~(TQ)~\cite{DBLP:conf/acl/JoshiCWZ17} and MS-MARCO Passage Ranking~(MS Pas)~\cite{nguyen2016ms}, a public document retrieval dataset: MS-MARCO Document Ranking~(MS Doc)~\cite{nguyen2016ms}, and an industry dataset that is collected from Bing search logs. Their statistics are shown in \cref{tab-dataset}.
The details of datasets, baselines and implementations are presented in Appendix.

\begin{table}[t]
\centering
\small
\begin{tabular}{l|cc}
\toprule 
\multicolumn{1}{l|}{\textbf{Method}} & \textbf{MRR@10} & \textbf{R@100} \\ \hline
BM25 & 0.279 & 0.807 \\ 
DPR~\citep{karpukhin2020dense} & 0.320 & 0.864 \\
ANCE~\citep{xiong2020approximate} & 0.377 & 0.894 \\
STAR~\citep{DBLP:conf/sigir/ZhanM0G0M21} & 0.390 & 0.913 \\
ADORE~\citep{DBLP:conf/sigir/ZhanM0G0M21} & 0.405 & \underline{0.919} \\
AR2~\citep{zhang2021adversarial}& \underline{0.418} & 0.914 \\
\hline
AR2+SimANS & \textbf{0.431} & \textbf{0.923} \\
\bottomrule
\end{tabular}
\caption{Performance on MS Doc development set. 
} 
\label{tab:doc}
\end{table}

\subsection{Results Analysis}
\paratitle{Performance on Public Retrieval Datasets.}
\cref{tab.mainresult} and \cref{tab:doc} show the experimental results on three public passage retrieval datasets.
First, we can see that AR2 outperforms most baseline methods on all datasets. AR2 incorporates an adversarial training framework to iteratively improve the retriever and ranker. 
Second, SimANS can further improve the performance of AR2, and outperform all baselines in terms of all the metrics across all datasets.
 SimANS only incorporates a new negative sampling strategy based on AR2, which aims to sample the ambiguous negatives that are neither too hard (potential false negatives) or too easy (uninformative).
According to the findings in \cref{sec-motivation}, 
such a way can alleviate the uninformative and false negative problems that are frequently encountered in commonly-used random and top-$k$ negatives sampling methods, and is able to sample high-quality negatives that contribute more to the model convergence.
Besides, the improvements of SimANS on AR2 are larger in MS Pas and Doc datasets than others.
The reason is that the two datasets are collected from real-world search logs that suffer severely from the false negative problem, whereas  SimANS is capable of alleviating this problem and provides better negatives for training.

\begin{table}[t]
\centering
\small
\begin{tabular}{l|ccc}
\toprule 
\multicolumn{1}{l|}{\textbf{Method}} & \textbf{R@5} & \textbf{R@20} & \textbf{R@100} \\ \hline
Baseline+Random Neg & 39.5 & 59.0 & 76.2 \\
Baseline+top-$k$ Neg & 57.1 & 73.5 & 85.1 \\ \hline
Baseline+SimANS & \textbf{59.1} & \textbf{74.9} & \textbf{85.6} \\
\bottomrule
\end{tabular}
\caption{Experimental results on Bing Industry dataset. } 
\label{tab:industry}
\end{table}

\paratitle{Performance on Bing Industry Dataset.}
For the Bing industry dataset, we adopt a dual-encoder mBERT~\cite{devlin2019bert} as the baseline model to deal with multilingual queries and documents, and implement different negative sampling strategies on it.
We simply evaluate the last checkpoint after training and report the results on the development set.
As shown in \cref{tab:industry}, after applying the top-$k$ hard negatives sampling, the performance of the baseline model is improved by a large margin. It indicates that hard negatives are more effective than randomly sampled ones.
Furthermore, we can see that  SimANS outperforms all other negative sampling methods, especially in Hit@5 (2\% absolute improvement).
It demonstrates the effectiveness of SimANS in industrial scenarios.
As a comparison, SimANS is able to alleviate the uninformative and false negatives problems that the random and top-$k$ negatives sampling strategies may suffer, respectively.

\begin{table}[t]
\centering
\small
\setlength\tabcolsep{5pt}
\begin{tabular}{l|cc|cc}
\toprule
\multicolumn{1}{l|}{\multirow{2}{*}{\textbf{Method}}} & \multicolumn{2}{c|}{\textbf{TQ}} & \multicolumn{2}{c}{\textbf{NQ}} \\
 & \textbf{R@5} & \multicolumn{1}{l|}{\textbf{R@20}} & \textbf{R@5} & \textbf{R@20} \\ \hline
ANCE & 72.4 & 80.3 & 71.8 & 81.9 \\
ANCE+SimANS & \textbf{74.8} & \textbf{82.1} & \textbf{74.3} & \textbf{83.0} \\ \hline 
RocketQA & 76.1 & 83.0 & 74.0 & 82.7 \\
RocketQA+SimANS & \textbf{77.1} & \textbf{83.6} & \textbf{76.7} & \textbf{84.8} \\
\bottomrule
\end{tabular}
\caption{The retrieval performance of applying our method on other baselines on TQ and NQ datasets} 
\label{tab:applying}
\end{table}

\subsection{Further Analysis}

\paratitle{Applying SimANS to Other Models.}
Since SimANS is a general negative sampling strategy, it can be applied to a variety of dense retrieval methods. Thus, in this part, we implement SimANS on two representative methods, ANCE~\cite{xiong2020approximate} and RocketQA~\cite{DBLP:conf/naacl/QuDLLRZDWW21}, as they adopt effective techniques as asynchronous index refresh and knowledge distillation, respectively.
We only replace the negative sampling strategies in these methods with SimANS and conduct experiments on TQ and NQ datasets. As shown in \cref{tab:applying}, our approach can consistently improve the performance of the two methods. It shows that SimANS is general to various dense retrieval methods with different techniques and can provide more high-quality negatives to improve their performance.

\begin{table}[t]
\centering
\small
\begin{tabular}{l|cccc}
\toprule
\textbf{Method} & \textbf{MRR@10} & \textbf{R@1} & \textbf{R@50} & \textbf{R@1k} \\ \hline
AR2 & 39.5 & 26.4 & 87.8 & 98.6 \\
AR2+Doc-Sim & 40.1 & 27.3 & 88.0 & 98.6 \\  
AR2+Nearest-K & 40.5 & 27.6 & 88.5 & \textbf{98.7} \\ \hline
AR2+SimANS & \textbf{40.9} & \textbf{28.2} & \textbf{88.7} & \textbf{98.7} \\
\bottomrule
\end{tabular}
\caption{The variation study of our method in AR2 on MS Pas development set.} 
\label{tab:variation}
\end{table}

\paratitle{Variation Study.}
Our proposed SimANS incorporates a new negative sampling probability distribution that is based on the differences between the query-document relevance scores of positives and negative candidates.
To verify the effectiveness of this distribution, we design two variations of SimANS: (1) \textbf{Doc-Sim} that leverages the document-document relevance scores between positives and negative candidates to replace the query-document relevance scores; (2) \textbf{Nearest-K} that directly picks the top-$k$ nearest negatives according to the differences of query-document relevance scores instead of sampling.
We implement these variations on AR2 and conduct experiments on the development set of MS Pas dataset. 
As shown in \cref{tab:variation},  SimANS outperforms all these variations. 
It indicates the effectiveness of our devised ambiguous negative sampling probability distribution.
For Doc-Sim, it is likely to select the false negatives that have similar semantics to positives, hurting the model performance.
For Nearest-K, as it always selects fixed negatives, it may cause overfitting.

\begin{figure}[t!]
    \centering
    \begin{subfigure}[b]{0.49\linewidth}
        \centering
        \includegraphics[width=\textwidth]{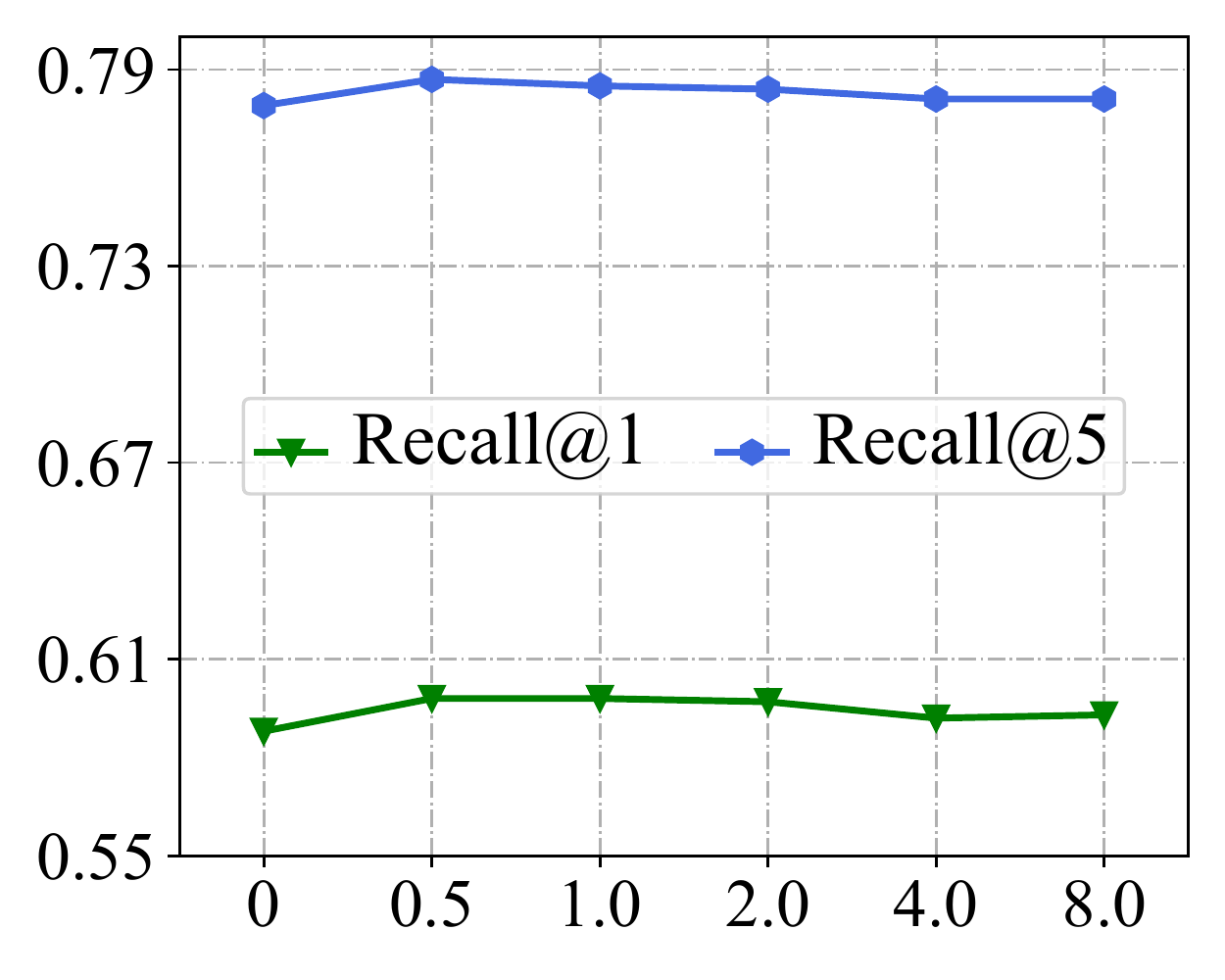}
        \caption{$a$: density hyper-parameter}
    \end{subfigure}
    \begin{subfigure}[b]{0.49\linewidth}
        \centering
        \includegraphics[width=\textwidth]{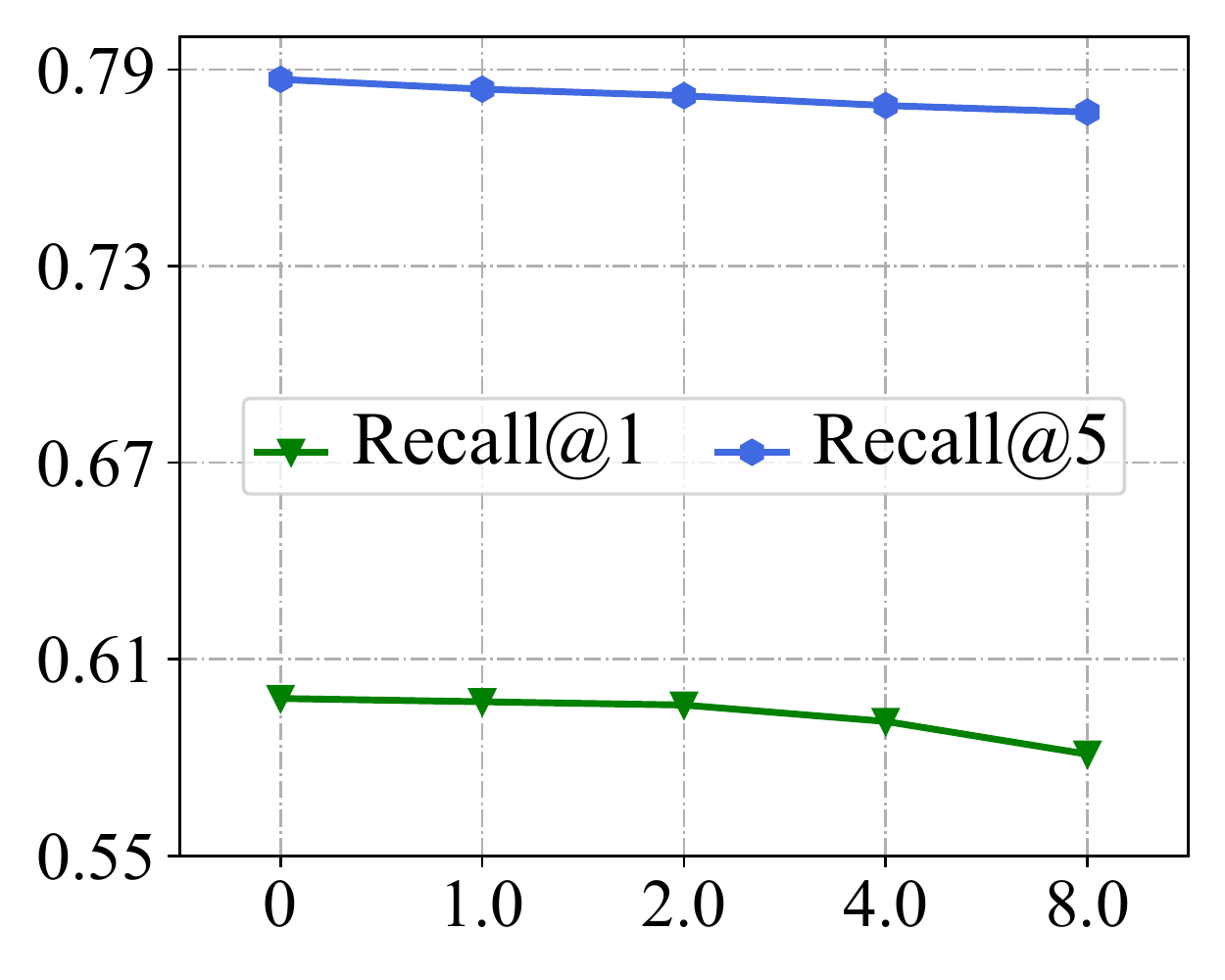}
        \caption{$b$: peak hyper-parameter}
    \end{subfigure}
    \caption{Performance comparison \emph{w.r.t.} hyper-parameters $a$ and $b$ on NQ dataset.}
    \vspace{-0.2cm}
\label{fig-param}
\end{figure}

\paratitle{Parameter Tuning.}
Our SimANS has two important hyper-parameters to tune, $a$ and $b$, which control the density and peak of the sampling probability distribution, respectively.
Here, we investigate the performance change of SimANS on AR2 \emph{w.r.t.} different $a$ and $b$ on NQ dataset.
As shown in \cref{fig-param}, our approach achieves the best performance when $a=0.5$ and $b=0$. It indicates that when the maximum point of the distribution has the same relevance score as the positive, the negative sampling probability distribution can produce more high-quality negatives.
Moreover, we notice that the model performance is not very sensitive to the two hyper-parameters if they are properly set within a certain range.

\begin{table}[t]
\centering
\small
\begin{tabular}{l|cc|cc}
\toprule
\multicolumn{1}{l|}{\multirow{2}{*}{\textbf{Ratio}}} & \multicolumn{2}{c|}{\textbf{AR2}} & \multicolumn{2}{c}{\textbf{AR2+SimANS}} \\
 & \textbf{R@5} & \multicolumn{1}{l|}{\textbf{Latency}} & \textbf{R@5} & \textbf{Latency} \\ \hline
$1:1$ & 76.4 & 210ms & 77.5 & 210ms \\
$1:5$ & 76.9 & 330ms & 78.1 & 340ms \\ 
$1:11$ & 77.1 & 510ms & 78.3 & 540ms \\
$1:15$ & 77.9 & 630ms & 78.7 & 650ms \\
\bottomrule
\end{tabular}
\caption{The retrieval performance and training latency \emph{w.r.t.} different sampled negative ratios on NQ dataset.} 
\label{tab:negative}
\end{table}

\paratitle{Impact of the Sampled Negative Ratio.}
We investigate the impact of the sampled negative ratio $1:k$ on retrieval performance and training latency per batch of SimANS on AR2. As shown in \cref{tab:negative}, with the increase of the sampled negative number, the performance improves consistently while the training latency  increases.
Besides, SimANS just slightly increases the training latency of AR2. It is because we can pre-compute the sampling probabilities before training, which avoids time-consuming computation during training.

\begin{figure}[t]
  \centering
  \includegraphics[width=\columnwidth]{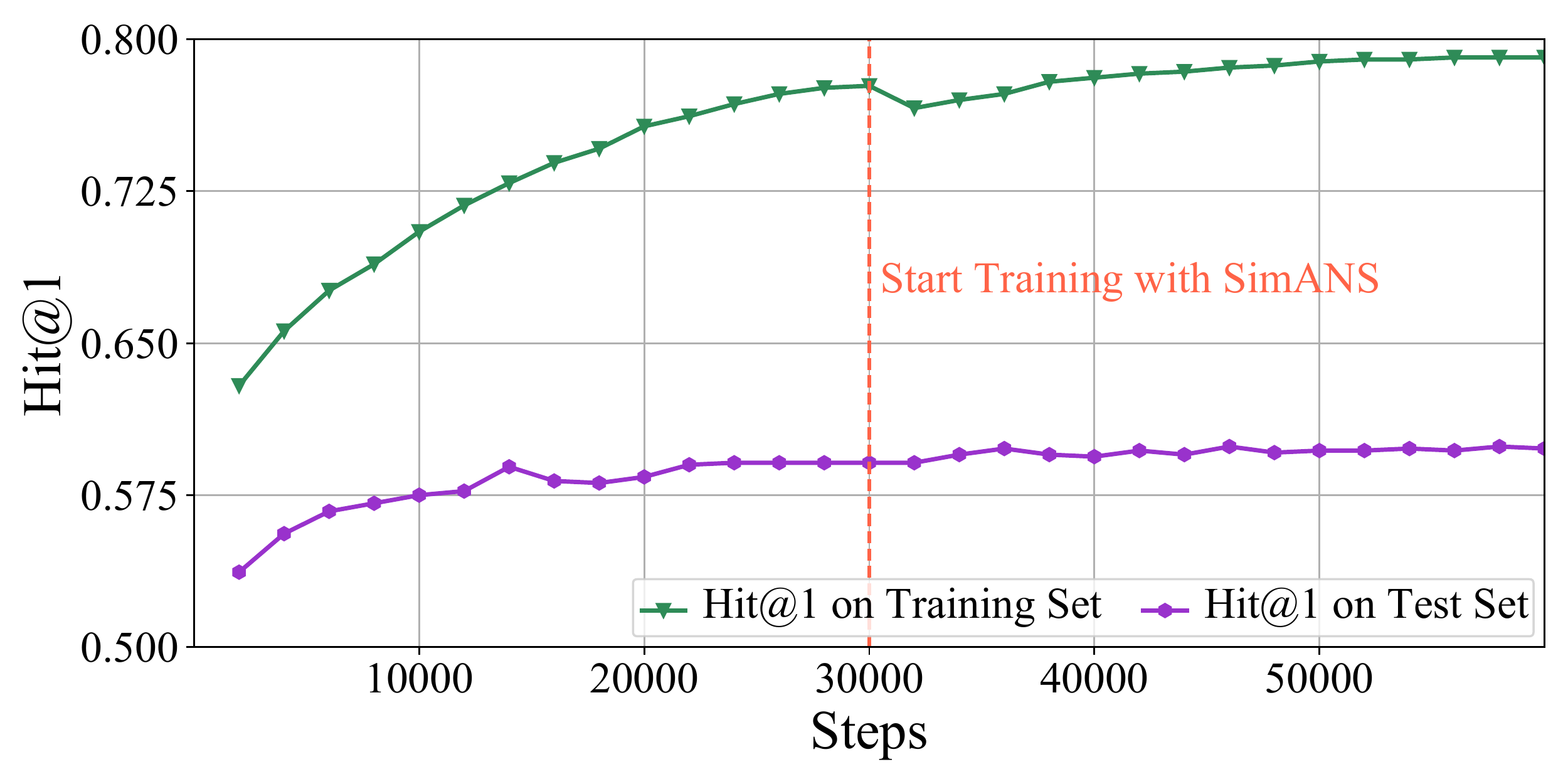}
  \caption{Hit@1 of AR2+SimANS on training and test sets of NQ \emph{w.r.t.} training steps.}
  \label{fig:training}
\end{figure}

\paratitle{Performance \emph{w.r.t.} Training Steps.}
Our approach requires continually training the model parameters that have been pre-trained by the original dense retrieval method. 
Here, we investigate the performance changes of the dense retrieval method before and after using SimANS \emph{w.r.t.} the training steps. We conduct experiments on AR2 and show the Hit@1 metric on NQ dataset in \cref{fig:training}. First, we can see that with the increase of the training steps, the performance of AR2 on training and test sets  improves simultaneously. 
After applying our SimANS, we can see that the performance further improves, especially in the training set ($0.777\longrightarrow0.791$). It indicates that our approach is capable of improving the fitting of the training set, and such an improvement can also generalize to the test set.

\section{Conclusion}

 We investigated how the gradient statistics of negative documents affect their relevance ranking for dense text retrieval. We discovered that negative documents with high gradient means and low gradient variances are more likely to be ambiguous negatives, which are informative and less prone to false negatives. Based on this insight, we proposed SimANS, a novel negative sampling method that balances the difficulty of negative examples by adjusting their sampling probabilities. SimANS improved the performance of various dense retrieval models on four public and one industrial datasets. We plan to apply our method to other information retrieval tasks, such as personal recommendation, and to develop better pre-training schemes for dense text retrieval in the future.

\section*{Acknowledgement}
Kun Zhou, Wayne Xin Zhao and Ji-Rong Wen are supported by Beijing Natural Science Foundation under Grant No. 4222027, and National Natural Science Foundation of China under Grant No. 61872369, Beijing Outstanding Young Scientist Program under Grant No. BJJWZYJH012019100020098, and the Outstanding Innovative Talents Cultivation Funded Programs 2021. Xin Zhao is the corresponding author.

\section*{Ethical Consideration}
In this section, we discuss the ethical considerations of this work from the following two aspects.
First, for intellectual property protection, the code, data and dense retrieval models adopted from previous works are granted for research-purpose usage.
Second, since PLMs have been shown to capture certain biases from the pre-trained corpus~\cite{bender2021dangers}, there is a potential problem about biases that are from the use of PLMs in our approach.
There are increasing efforts to address this problem in the community~\cite{ross2020measuring}.

\bibliography{anthology,custom}

\begin{thebibliography}{48}
\expandafter\ifx\csname natexlab\endcsname\relax\def\natexlab#1{#1}\fi

\bibitem[{Bender et~al.(2021)Bender, Gebru, McMillan-Major, and
  Shmitchell}]{bender2021dangers}
Emily~M Bender, Timnit Gebru, Angelina McMillan-Major, and Shmargaret
  Shmitchell. 2021.
\newblock On the dangers of stochastic parrots: Can language models be too big?
\newblock In \emph{Proceedings of the 2021 ACM Conference on Fairness,
  Accountability, and Transparency}, pages 610--623.

\bibitem[{Brickley et~al.(2019)Brickley, Burgess, and Noy}]{brickley2019google}
Dan Brickley, Matthew Burgess, and Natasha Noy. 2019.
\newblock Google dataset search: Building a search engine for datasets in an
  open web ecosystem.
\newblock In \emph{WWW}.

\bibitem[{Dai and Callan(2019)}]{dai2019deeper}
Zhuyun Dai and Jamie Callan. 2019.
\newblock Deeper text understanding for ir with contextual neural language
  modeling.
\newblock In \emph{SIGIR}.

\bibitem[{Devlin et~al.(2019)Devlin, Chang, Lee, and
  Toutanova}]{devlin2019bert}
Jacob Devlin, Ming-Wei Chang, Kenton Lee, and Kristina Toutanova. 2019.
\newblock Bert: Pre-training of deep bidirectional transformers for language
  understanding.
\newblock In \emph{NAACL}.

\bibitem[{Gao and Callan(2021)}]{gao2021your}
Luyu Gao and Jamie Callan. 2021.
\newblock Is your language model ready for dense representation fine-tuning?
\newblock \emph{arXiv preprint arXiv:2104.08253}.

\bibitem[{Gao and Callan(2022)}]{gao2021unsupervised}
Luyu Gao and Jamie Callan. 2022.
\newblock Unsupervised corpus aware language model pre-training for dense
  passage retrieval.
\newblock In \emph{ACL}, pages 2843--2853.

\bibitem[{Gao et~al.(2021{\natexlab{a}})Gao, Dai, and
  Callan}]{DBLP:conf/naacl/GaoDC21}
Luyu Gao, Zhuyun Dai, and Jamie Callan. 2021{\natexlab{a}}.
\newblock {COIL:} revisit exact lexical match in information retrieval with
  contextualized inverted list.
\newblock In \emph{NAACL-HLT}.

\bibitem[{Gao et~al.(2021{\natexlab{b}})Gao, Zhang, Han, and
  Callan}]{gao-etal-2021-scaling}
Luyu Gao, Yunyi Zhang, Jiawei Han, and Jamie Callan. 2021{\natexlab{b}}.
\newblock Scaling deep contrastive learning batch size under memory limited
  setup.
\newblock In \emph{Proceedings of the 6th Workshop on Representation Learning
  for NLP (RepL4NLP-2021)}.

\bibitem[{Hofst{\"a}tter et~al.(2021)Hofst{\"a}tter, Lin, Yang, Lin, and
  Hanbury}]{hofstatter2021efficiently}
Sebastian Hofst{\"a}tter, Sheng-Chieh Lin, Jheng-Hong Yang, Jimmy Lin, and
  Allan Hanbury. 2021.
\newblock Efficiently teaching an effective dense retriever with balanced topic
  aware sampling.
\newblock In \emph{Proceedings of the 44th International ACM SIGIR Conference
  on Research and Development in Information Retrieval}, pages 113--122.

\bibitem[{Hong et~al.(2022)Hong, Zhang, Wang, and Zhao}]{hong2022sentence}
Wu~Hong, Zhuosheng Zhang, Jinyuan Wang, and Hai Zhao. 2022.
\newblock Sentence-aware contrastive learning for open-domain passage
  retrieval.
\newblock In \emph{ACL}, pages 1062--1074.

\bibitem[{Izacard and Grave(2021)}]{izacard2021leveraging}
Gautier Izacard and {\'E}douard Grave. 2021.
\newblock Leveraging passage retrieval with generative models for open domain
  question answering.
\newblock In \emph{Proceedings of the 16th Conference of the European Chapter
  of the Association for Computational Linguistics: Main Volume}, pages
  874--880.

\bibitem[{Johnson et~al.(2019)Johnson, Douze, and
  J{\'e}gou}]{johnson2019billion}
Jeff Johnson, Matthijs Douze, and Herv{\'e} J{\'e}gou. 2019.
\newblock Billion-scale similarity search with gpus.
\newblock \emph{IEEE Transactions on Big Data}.

\bibitem[{Johnson and Guestrin(2018)}]{johnson2018training}
Tyler~B Johnson and Carlos Guestrin. 2018.
\newblock Training deep models faster with robust, approximate importance
  sampling.
\newblock \emph{Advances in Neural Information Processing Systems}, 31.

\bibitem[{Joshi et~al.(2017)Joshi, Choi, Weld, and
  Zettlemoyer}]{DBLP:conf/acl/JoshiCWZ17}
Mandar Joshi, Eunsol Choi, Daniel~S. Weld, and Luke Zettlemoyer. 2017.
\newblock Triviaqa: {A} large scale distantly supervised challenge dataset for
  reading comprehension.
\newblock In \emph{ACL}.

\bibitem[{Karpukhin et~al.(2020)Karpukhin, Oguz, Min, Lewis, Wu, Edunov, Chen,
  and Yih}]{karpukhin2020dense}
Vladimir Karpukhin, Barlas Oguz, Sewon Min, Patrick S.~H. Lewis, Ledell Wu,
  Sergey Edunov, Danqi Chen, and Wen{-}tau Yih. 2020.
\newblock Dense passage retrieval for open-domain question answering.
\newblock In \emph{EMNLP}.

\bibitem[{Katharopoulos and Fleuret(2018)}]{katharopoulos2018not}
Angelos Katharopoulos and Fran{\c{c}}ois Fleuret. 2018.
\newblock Not all samples are created equal: Deep learning with importance
  sampling.
\newblock In \emph{International conference on machine learning}, pages
  2525--2534. PMLR.

\bibitem[{Koh and Liang(2017)}]{koh2017understanding}
Pang~Wei Koh and Percy Liang. 2017.
\newblock Understanding black-box predictions via influence functions.
\newblock In \emph{International conference on machine learning}, pages
  1885--1894. PMLR.

\bibitem[{Kwiatkowski et~al.(2019)Kwiatkowski, Palomaki, Redfield, Collins,
  Parikh, Alberti, Epstein, Polosukhin, Devlin, Lee, Toutanova, Jones, Kelcey,
  Chang, Dai, Uszkoreit, Le, and Petrov}]{DBLP:journals/tacl/KwiatkowskiPRCP19}
Tom Kwiatkowski, Jennimaria Palomaki, Olivia Redfield, Michael Collins,
  Ankur~P. Parikh, Chris Alberti, Danielle Epstein, Illia Polosukhin, Jacob
  Devlin, Kenton Lee, Kristina Toutanova, Llion Jones, Matthew Kelcey,
  Ming{-}Wei Chang, Andrew~M. Dai, Jakob Uszkoreit, Quoc Le, and Slav Petrov.
  2019.
\newblock Natural questions: a benchmark for question answering research.
\newblock \emph{Trans. Assoc. Comput. Linguistics}, 7:452--466.

\bibitem[{Lu et~al.(2022)Lu, Liu, Liu, Shi, Huang, Sun, Tian, Wu, Wang, Yin
  et~al.}]{lu2022ernie}
Yuxiang Lu, Yiding Liu, Jiaxiang Liu, Yunsheng Shi, Zhengjie Huang, Shikun
  Feng~Yu Sun, Hao Tian, Hua Wu, Shuaiqiang Wang, Dawei Yin, et~al. 2022.
\newblock Ernie-search: Bridging cross-encoder with dual-encoder via self
  on-the-fly distillation for dense passage retrieval.
\newblock \emph{arXiv preprint arXiv:2205.09153}.

\bibitem[{Luan et~al.(2021)Luan, Eisenstein, Toutanova, and
  Collins}]{luan2021sparse}
Yi~Luan, Jacob Eisenstein, Kristina Toutanova, and Michael Collins. 2021.
\newblock Sparse, dense, and attentional representations for text retrieval.
\newblock \emph{Transactions of the Association for Computational Linguistics},
  9:329--345.

\bibitem[{Mao et~al.(2022)Mao, Dou, and Qian}]{mao2022curriculum}
Kelong Mao, Zhicheng Dou, and Hongjin Qian. 2022.
\newblock Curriculum contrastive context denoising for few-shot conversational
  dense retrieval.
\newblock In \emph{SIGIR}, pages 176--186.

\bibitem[{Mao et~al.(2021)Mao, He, Liu, Shen, Gao, Han, and
  Chen}]{DBLP:conf/acl/MaoHLSG0C20}
Yuning Mao, Pengcheng He, Xiaodong Liu, Yelong Shen, Jianfeng Gao, Jiawei Han,
  and Weizhu Chen. 2021.
\newblock Generation-augmented retrieval for open-domain question answering.
\newblock In \emph{ACL}.

\bibitem[{Meissner et~al.(2021)Meissner, Thumwanit, Sugawara, and
  Aizawa}]{meissner2021embracing}
Johannes~Mario Meissner, Napat Thumwanit, Saku Sugawara, and Akiko Aizawa.
  2021.
\newblock Embracing ambiguity: Shifting the training target of nli models.
\newblock In \emph{ACL}, pages 862--869.

\bibitem[{Min et~al.(2020)Min, Michael, Hajishirzi, and
  Zettlemoyer}]{min2020ambigqa}
Sewon Min, Julian Michael, Hannaneh Hajishirzi, and Luke Zettlemoyer. 2020.
\newblock Ambigqa: Answering ambiguous open-domain questions.
\newblock In \emph{EMNLP}, pages 5783--5797.

\bibitem[{Nguyen et~al.(2016)Nguyen, Rosenberg, Song, Gao, Tiwary, Majumder,
  and Deng}]{nguyen2016ms}
Tri Nguyen, Mir Rosenberg, Xia Song, Jianfeng Gao, Saurabh Tiwary, Rangan
  Majumder, and Li~Deng. 2016.
\newblock Ms marco: A human generated machine reading comprehension dataset.
\newblock In \emph{CoCo@ NIPS}.

\bibitem[{Nogueira et~al.(2019{\natexlab{a}})Nogueira, Lin, and
  Epistemic}]{nogueira2019doc2query}
Rodrigo Nogueira, Jimmy Lin, and AI~Epistemic. 2019{\natexlab{a}}.
\newblock From doc2query to doctttttquery.
\newblock \emph{Online preprint}.

\bibitem[{Nogueira et~al.(2019{\natexlab{b}})Nogueira, Yang, Lin, and
  Cho}]{nogueira2019document}
Rodrigo Nogueira, Wei Yang, Jimmy Lin, and Kyunghyun Cho. 2019{\natexlab{b}}.
\newblock Document expansion by query prediction.
\newblock \emph{arXiv preprint arXiv:1904.08375}.

\bibitem[{O{\u{g}}uz et~al.(2022)O{\u{g}}uz, Lakhotia, Gupta, Lewis, Karpukhin,
  Piktus, Chen, Riedel, Yih, Gupta et~al.}]{ouguz2021domain}
Barlas O{\u{g}}uz, Kushal Lakhotia, Anchit Gupta, Patrick Lewis, Vladimir
  Karpukhin, Aleksandra Piktus, Xilun Chen, Sebastian Riedel, Wen-tau Yih,
  Sonal Gupta, et~al. 2022.
\newblock Domain-matched pre-training tasks for dense retrieval.
\newblock In \emph{Findings of NAACL}.

\bibitem[{Pruthi et~al.(2020)Pruthi, Liu, Kale, and
  Sundararajan}]{pruthi2020estimating}
Garima Pruthi, Frederick Liu, Satyen Kale, and Mukund Sundararajan. 2020.
\newblock Estimating training data influence by tracing gradient descent.
\newblock \emph{Advances in Neural Information Processing Systems},
  33:19920--19930.

\bibitem[{Qiu et~al.(2022)Qiu, Li, Qu, Chen, She, Liu, Wu, and
  Wang}]{qiu2022dureader_retrieval}
Yifu Qiu, Hongyu Li, Yingqi Qu, Ying Chen, Qiaoqiao She, Jing Liu, Hua Wu, and
  Haifeng Wang. 2022.
\newblock Dureader\_retrieval: A large-scale chinese benchmark for passage
  retrieval from web search engine.
\newblock \emph{arXiv preprint arXiv:2203.10232}.

\bibitem[{Qu et~al.(2021)Qu, Ding, Liu, Liu, Ren, Zhao, Dong, Wu, and
  Wang}]{DBLP:conf/naacl/QuDLLRZDWW21}
Yingqi Qu, Yuchen Ding, Jing Liu, Kai Liu, Ruiyang Ren, Wayne~Xin Zhao, Daxiang
  Dong, Hua Wu, and Haifeng Wang. 2021.
\newblock Rocketqa: An optimized training approach to dense passage retrieval
  for open-domain question answering.
\newblock In \emph{NAACL-HLT}.

\bibitem[{Ram et~al.(2022)Ram, Shachaf, Levy, Berant, and
  Globerson}]{ram2021learning}
Ori Ram, Gal Shachaf, Omer Levy, Jonathan Berant, and Amir Globerson. 2022.
\newblock Learning to retrieve passages without supervision.
\newblock In \emph{NAACL}.

\bibitem[{Reimers and Gurevych(2019)}]{reimers2019sentence}
Nils Reimers and Iryna Gurevych. 2019.
\newblock Sentence-bert: Sentence embeddings using siamese bert-networks.
\newblock In \emph{Proceedings of the 2019 Conference on Empirical Methods in
  Natural Language Processing and the 9th International Joint Conference on
  Natural Language Processing (EMNLP-IJCNLP)}, pages 3982--3992.

\bibitem[{Ren et~al.(2021{\natexlab{a}})Ren, Lv, Qu, Liu, Zhao, She, Wu, Wang,
  and Wen}]{DBLP:conf/acl/RenLQLZSWWW21}
Ruiyang Ren, Shangwen Lv, Yingqi Qu, Jing Liu, Wayne~Xin Zhao, Qiaoqiao She,
  Hua Wu, Haifeng Wang, and Ji{-}Rong Wen. 2021{\natexlab{a}}.
\newblock {PAIR:} leveraging passage-centric similarity relation for improving
  dense passage retrieval.
\newblock In \emph{Findings of {ACL/IJCNLP}}.

\bibitem[{Ren et~al.(2021{\natexlab{b}})Ren, Qu, Liu, Zhao, She, Wu, Wang, and
  Wen}]{ren2021rocketqav2}
Ruiyang Ren, Yingqi Qu, Jing Liu, Wayne~Xin Zhao, Qiaoqiao She, Hua Wu, Haifeng
  Wang, and Ji-Rong Wen. 2021{\natexlab{b}}.
\newblock Rocketqav2: A joint training method for dense passage retrieval and
  passage re-ranking.
\newblock In \emph{EMNLP}, pages 2825--2835.

\bibitem[{Ross et~al.(2021)Ross, Katz, and Barbu}]{ross2020measuring}
Candace Ross, Boris Katz, and Andrei Barbu. 2021.
\newblock Measuring social biases in grounded vision and language embeddings.
\newblock In \emph{NAACL}, pages 998--1008.

\bibitem[{Sachan et~al.(2021)Sachan, Patwary, Shoeybi, Kant, Ping, Hamilton,
  and Catanzaro}]{sachan2021end}
Devendra~Singh Sachan, Mostofa Patwary, Mohammad Shoeybi, Neel Kant, Wei Ping,
  William~L. Hamilton, and Bryan Catanzaro. 2021.
\newblock End-to-end training of neural retrievers for open-domain question
  answering.
\newblock In \emph{ACL/IJCNLP}.

\bibitem[{Sun et~al.(2020)Sun, Wang, Li, Feng, Tian, Wu, and
  Wang}]{sun2020ernie}
Yu~Sun, Shuohuan Wang, Yukun Li, Shikun Feng, Hao Tian, Hua Wu, and Haifeng
  Wang. 2020.
\newblock Ernie 2.0: A continual pre-training framework for language
  understanding.
\newblock In \emph{AAAI}.

\bibitem[{Swayamdipta et~al.(2020)Swayamdipta, Schwartz, Lourie, Wang,
  Hajishirzi, Smith, and Choi}]{swayamdipta2020dataset}
Swabha Swayamdipta, Roy Schwartz, Nicholas Lourie, Yizhong Wang, Hannaneh
  Hajishirzi, Noah~A Smith, and Yejin Choi. 2020.
\newblock Dataset cartography: Mapping and diagnosing datasets with training
  dynamics.
\newblock In \emph{EMNLP}, pages 9275--9293.

\bibitem[{Xiong et~al.(2021)Xiong, Xiong, Li, Tang, Liu, Bennett, Ahmed, and
  Overwijk}]{xiong2020approximate}
Lee Xiong, Chenyan Xiong, Ye~Li, Kwok{-}Fung Tang, Jialin Liu, Paul~N. Bennett,
  Junaid Ahmed, and Arnold Overwijk. 2021.
\newblock Approximate nearest neighbor negative contrastive learning for dense
  text retrieval.
\newblock In \emph{ICLR}.

\bibitem[{Xu et~al.(2022)Xu, Guo, Duan, and McAuley}]{xu2022laprador}
Canwen Xu, Daya Guo, Nan Duan, and Julian McAuley. 2022.
\newblock Laprador: Unsupervised pretrained dense retriever for zero-shot text
  retrieval.
\newblock In \emph{Findings of ACL}, pages 3557--3569.

\bibitem[{Yang et~al.(2017)Yang, Fang, and Lin}]{yang2017anserini}
Peilin Yang, Hui Fang, and Jimmy Lin. 2017.
\newblock Anserini: Enabling the use of lucene for information retrieval
  research.
\newblock In \emph{SIGIR}.

\bibitem[{Yang and Seo(2020)}]{yang2020retriever}
Sohee Yang and Minjoon Seo. 2020.
\newblock Is retriever merely an approximator of reader?
\newblock \emph{arXiv preprint arXiv:2010.10999}.

\bibitem[{Zhan et~al.(2021)Zhan, Mao, Liu, Guo, Zhang, and
  Ma}]{DBLP:conf/sigir/ZhanM0G0M21}
Jingtao Zhan, Jiaxin Mao, Yiqun Liu, Jiafeng Guo, Min Zhang, and Shaoping Ma.
  2021.
\newblock Optimizing dense retrieval model training with hard negatives.
\newblock In \emph{SIGIR}.

\bibitem[{Zhan et~al.(2020)Zhan, Mao, Liu, Zhang, and Ma}]{zhan2020repbert}
Jingtao Zhan, Jiaxin Mao, Yiqun Liu, Min Zhang, and Shaoping Ma. 2020.
\newblock Repbert: Contextualized text embeddings for first-stage retrieval.
\newblock \emph{arXiv preprint arXiv:2006.15498}.

\bibitem[{Zhang et~al.(2021)Zhang, Gong, Shen, Lv, Duan, and
  Chen}]{zhang2021adversarial}
Hang Zhang, Yeyun Gong, Yelong Shen, Jiancheng Lv, Nan Duan, and Weizhu Chen.
  2021.
\newblock Adversarial retriever-ranker for dense text retrieval.
\newblock In \emph{International Conference on Learning Representations}.

\bibitem[{Zhou et~al.(2022{\natexlab{a}})Zhou, Li, Shang, Luo, Zhan, Hu, Zhang,
  Jiang, Cao, Yu et~al.}]{zhou2022hyperlink}
Jiawei Zhou, Xiaoguang Li, Lifeng Shang, Lan Luo, Ke~Zhan, Enrui Hu, Xinyu
  Zhang, Hao Jiang, Zhao Cao, Fan Yu, et~al. 2022{\natexlab{a}}.
\newblock Hyperlink-induced pre-training for passage retrieval in open-domain
  question answering.
\newblock In \emph{ACL}, pages 7135--7146.

\bibitem[{Zhou et~al.(2022{\natexlab{b}})Zhou, Zhang, Zhao, and
  Wen}]{zhou2022debiased}
Kun Zhou, Beichen Zhang, Wayne~Xin Zhao, and Ji-Rong Wen. 2022{\natexlab{b}}.
\newblock Debiased contrastive learning of unsupervised sentence
  representations.
\newblock In \emph{Proceedings of the 60th Annual Meeting of the Association
  for Computational Linguistics (Volume 1: Long Papers)}, pages 6120--6130.

\end{thebibliography}
\bibliographystyle{acl_natbib}
\newpage
\appendix

\section{Illustration of Ambiguous Negatives}

\begin{figure}[t]
  \centering
  \includegraphics[width=0.9\columnwidth]{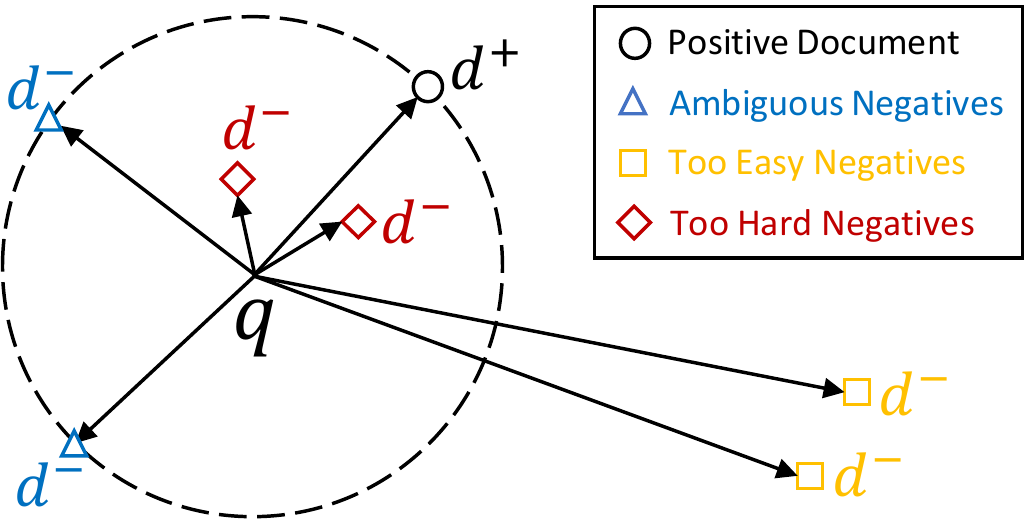}
  \caption{An example of the dense embedding distribution of a query with its positive document, too easy, too hard and ambiguous negatives.}
  \label{fig:case}
\end{figure}

We illustrate the distribution of the dense embeddings of a query with its positive document, too easy, too hard and ambiguous negatives in \cref{fig:case}.
Too hard negatives have a higher risk of being false negatives, and we can see that their dense embeddings locate closely to the ones of the query and the positive. If we learn to push them away, the distances between the embeddings of the query and the positive may also be enlarged, which is harmful to the goal of pulling the query and its positives together. 
Besides, too easy negatives locate rather far from the query, hence it is unnecessary to learn to push them even further.
As a comparison, the ambiguous negatives have similar distances as the positive, which compose the \emph{circular boundary} for the document pool consisting of hard negatives required to learn (\ie push away).
In this way, our SimANS can be seen as always sampling the borderline hard negatives from the document pool. By learning to push them away, we can narrow the circular boundary of hard negatives, which helps gradually achieve the goal that pulls the query and positives together while pushing apart negatives.

\section{More Details on Datasets}
We conduct experiments on five datasets, consisting of three passage retrieval datasets: Natural Question~(NQ)~\cite{DBLP:journals/tacl/KwiatkowskiPRCP19}, Trivia QA~(TQ)~\cite{DBLP:conf/acl/JoshiCWZ17} and MS-MARCO Passage Ranking~(MS Pas)~\cite{nguyen2016ms}, a document retrieval dataset: MS-MARCO Document Ranking~(MS Doc)~\cite{nguyen2016ms} and a real-world industry dataset Bing.
NQ and TQ are open domain question answering datasets collected from Google search logs and authored by trivia enthusiasts, respectively.
In the two datasets, each question is paired with an answer span and several golden passages from Wikipedia articles.
Following existing works~\cite{zhang2021adversarial,sachan2021end}, we adopt Recall@k (R@k) as the evaluation metrics, which measures if the top-$k$ ranked documents include the answer span. 
MS Pas and MS Doc consist of real questions collected from Bing search logs, where each question is paired with several web passages and documents, respectively.
As their labels of test sets are not available, we follow existing works~\cite{ren2021rocketqav2,DBLP:conf/sigir/ZhanM0G0M21} that report results on their development sets and adopt MRR@10, R@50 and R@1k for MS Pas, MRR@10 and R@100 for MS Doc.
Bing is collected from Bing search logs, where each example consists of a user historical query and several documents that the user has clicked.
These documents are real-world webpages and may contain hyperlinks and different languages.
We select Hit@5, Hit@20 and Hit@100 for evaluation.

\section{More Details on Baselines}
We compare our approach with a variety of methods, including sparse and dense retrieval models.

$\bullet$ \textbf{BM25}~\cite{yang2017anserini} is a widely-used sparse retriever based on exact matching.

$\bullet$ \textbf{GAR}~\cite{DBLP:conf/acl/MaoHLSG0C20}, \textbf{doc2query}~\cite{nogueira2019doc2query}, \textbf{DeepCT}~\cite{dai2019deeper} and \textbf{docTTTTTquery}~\cite{nogueira2019document} enhance BM25 by incorporating neural models.

$\bullet$ \textbf{DPR}~\cite{karpukhin2020dense}, \textbf{ANCE}~\cite{xiong2020approximate} and \textbf{STAR}~\cite{DBLP:conf/sigir/ZhanM0G0M21} are dense retrieval methods that adopt top-$k$ hard negatives to improve training.

$\bullet$ \textbf{COIL}~\cite{gao-etal-2021-scaling} and \textbf{ME-BERT}~\cite{luan2021sparse} combine sparse and dense representations for text retrieval.

$\bullet$ \textbf{Joint and Individual top-$k$}~\cite{sachan2021end} propose to train the dense retrieval model in an end-to-end manner.

$\bullet$ \textbf{RocketQA}~\cite{DBLP:conf/naacl/QuDLLRZDWW21}, \textbf{RDR}~\cite{yang2020retriever}, \textbf{RocketQAv2}~\cite{ren2021rocketqav2} and \textbf{ERNIE-search}~\cite{lu2022ernie} utilize knowledge distillation technique that leverages a teacher model to guide the training of the dense retrieval model.

$\bullet$ \textbf{PAIR}~\cite{DBLP:conf/acl/RenLQLZSWWW21}, \textbf{DPR-PAQ}~\cite{ouguz2021domain}, \textbf{Condenser}~\cite{gao2021your} and \textbf{coCondenser}~\cite{gao2021unsupervised} design special pre-training tasks to improve the backbone model for the dense retrieval task.

$\bullet$ \textbf{AR2}~\cite{zhang2021adversarial} incorporates an adversarial framework to jointly train the retriever and the ranker.
As it has achieved state-of-the-art performance on most datasets, we implement our approach on it to verify its effectiveness.

\section{Experimental Details}
\paratitle{Implementation Details on Public Datasets.}
For three passage retrieval tasks, we follow the experimental settings in AR2~\cite{zhang2021adversarial} that selects ERNIE-2.0-base~\cite{sun2020ernie} as the backbone model.
For MS Doc dataset, we leverage the model parameters of STAR~\cite{DBLP:conf/sigir/ZhanM0G0M21} to initialize AR2, and then train AR2 with the same hyper-parameters as STAR until convergence.
Next, we continue to train the AR2 model parameters with our proposed SimANS, where we set $a$ and $b$ to \{(0.5, 1.0), (0.5, 0) , (0.5, 0) , (0.5, 0)\} for NQ, TQ, MS Pas and MS Doc datasets, respectively.
The learning rate is set to 1e-5 for NQ and 5e-6 for other datasets.
The batch size is 256 for MS-Pas and MS-Doc, 64 for NQ and TQ, and the sampling ratio of positives and negatives is 1:15. All other hyper-parameter settings are the same as AR2.
All the experiments in this work are conducted on 8 NVIDIA Tesla A100 GPUs.

\paratitle{Implementation Details on Bing Industry Dataset.}
For the industry dataset, Bing, we adopt mBERT-base~\cite{devlin2019bert} as the backbone of the query and document encoders, to deal with multilingual queries and documents.
The parameters of the baseline model are trained with randomly sampled negatives using the infoNCE loss~\cite{karpukhin2020dense}, namely \textbf{Baseline+Random Neg}, and the sampling ratio of positives and negatives is 1:5.
The learning rate is 1e-5, the batch size is 128 and the training step is 100,000.
As a comparison, we implement the top-$k$ negatives sampling strategy on the baseline model, namely \textbf{Baseline+top-$k$ Neg}, where we utilize the baseline model to rank and select the top 5 documents that do not contain the query as hard negatives.
In our approach, namely \textbf{Baseline+SimANS}, we continue to train the Baseline+top-$k$ Neg model, but apply our SimANS to sample 5 negatives from the top 100 ranked documents.
We set $a$ to 1, $b$ to 0, and reuse the other hyper-parameters of the Baseline+top-$k$ Neg model.

\begin{figure*}[t!]
    \centering
    \begin{subfigure}[b]{0.49\linewidth}
        \centering
        \includegraphics[width=\textwidth]{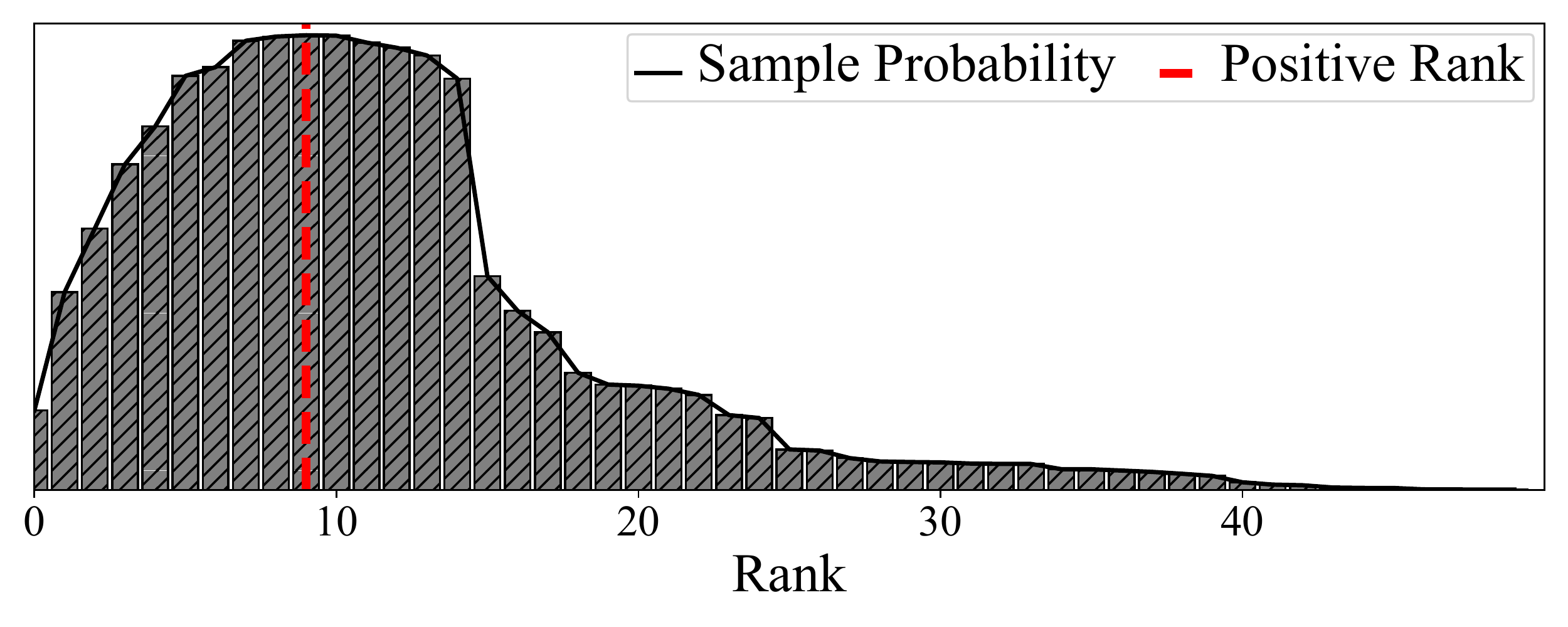}
        \caption{}
        \label{case-a}
    \end{subfigure}
    \begin{subfigure}[b]{0.49\linewidth}
        \centering
        \includegraphics[width=\textwidth]{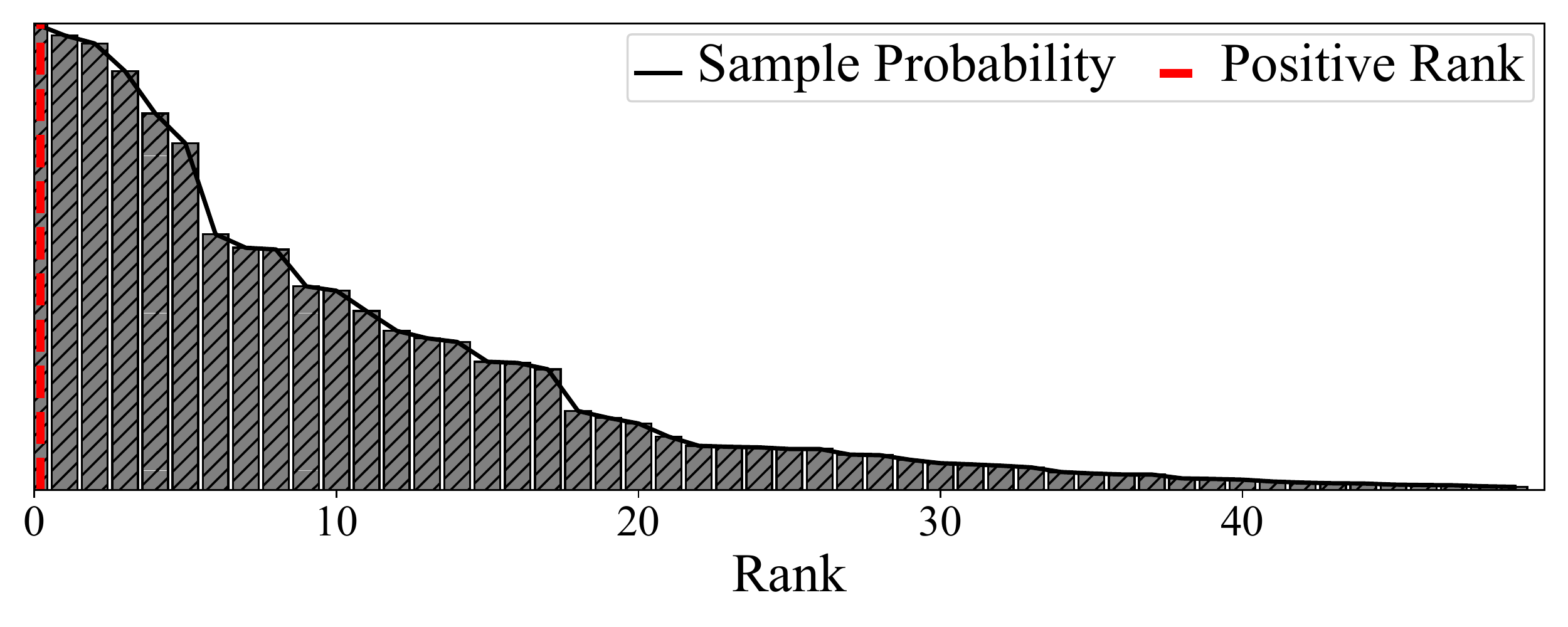}
        \caption{}
        \label{case-b}
    \end{subfigure}
    \quad
    \begin{subfigure}[b]{0.49\linewidth}
        \centering
        \includegraphics[width=\textwidth]{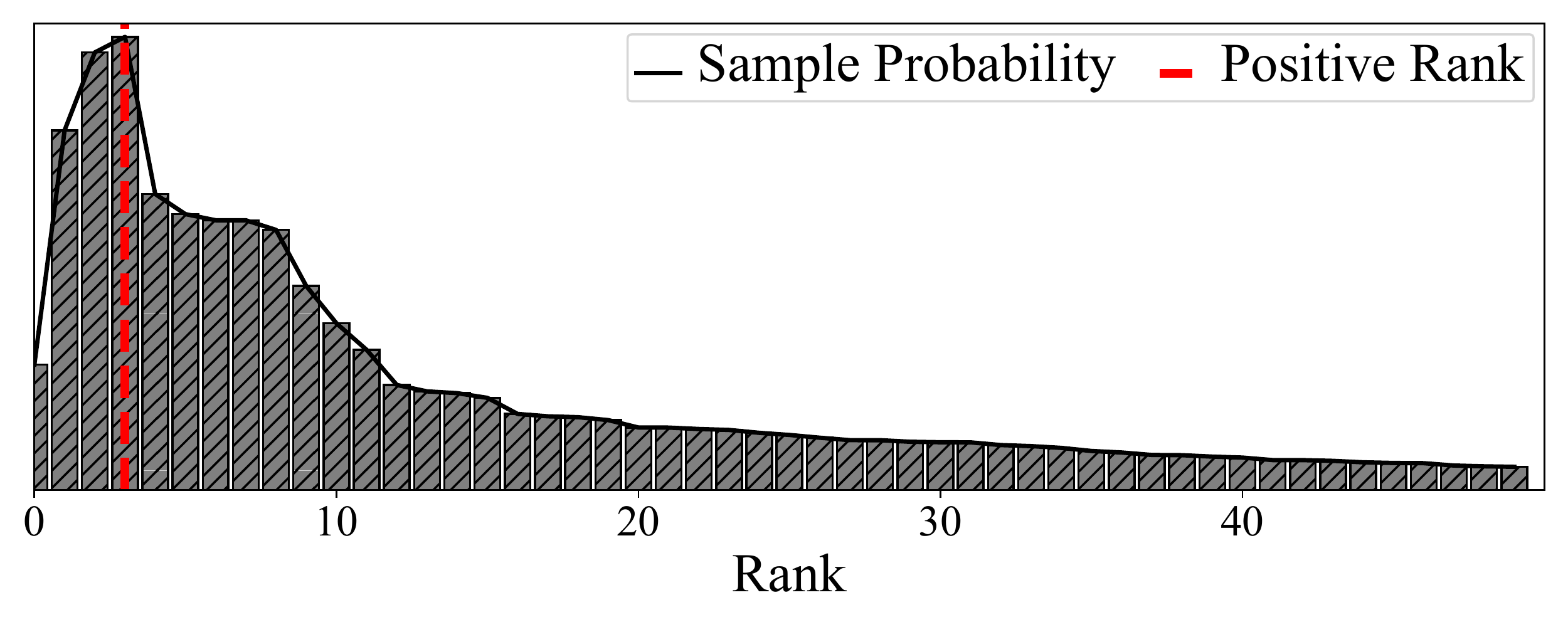}
        \caption{}
        \label{case-c}
    \end{subfigure}
    \begin{subfigure}[b]{0.49\linewidth}
        \centering
        \includegraphics[width=\textwidth]{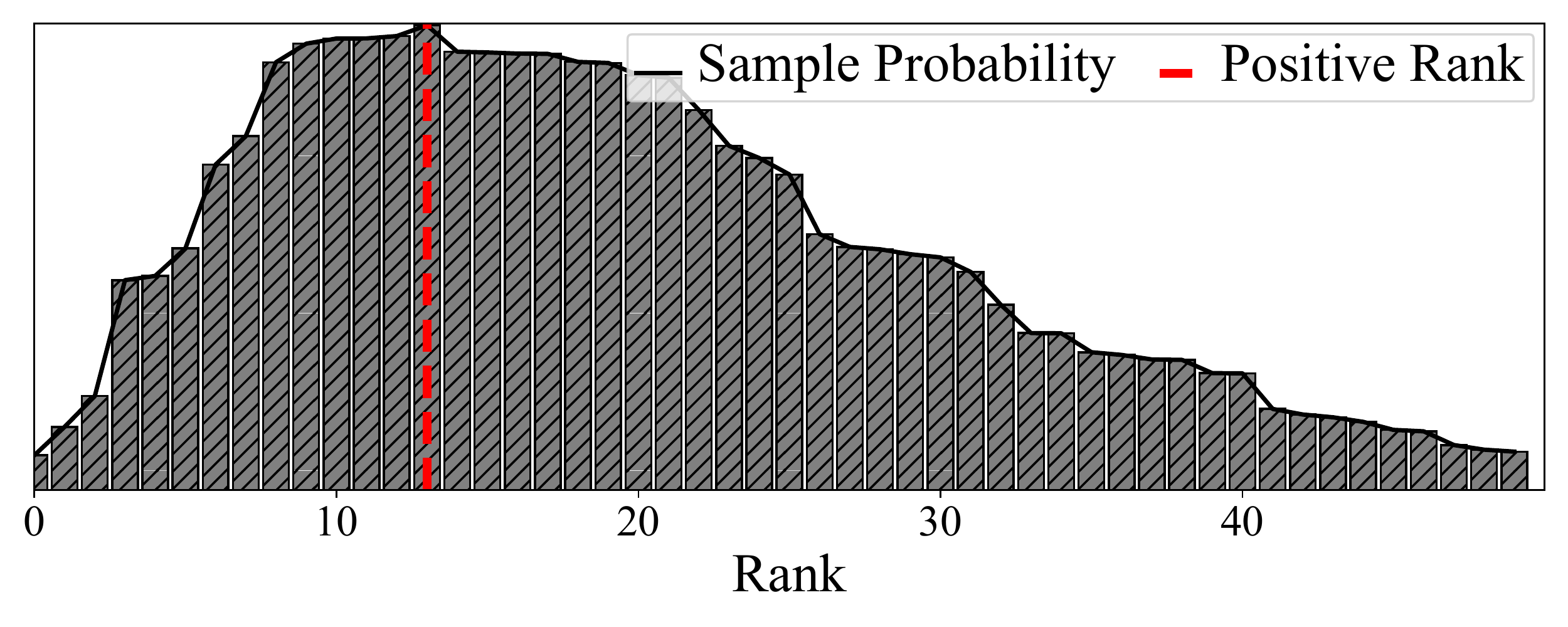}
        \caption{}
        \label{case-d}
    \end{subfigure}
    \caption{Illustration of four sampling probability distributions of the top 50 ranked negatives generated by our SimANS on the training set of MS Pas.}
    \vspace{-0.2cm}
\label{fig:case_study}
\end{figure*}

\section{Case Study}
In this part, we show four examples of the generated sampling probability distributions by our SimANS. These four examples are randomly selected from the training set of MS Pas dataset.
As shown in \cref{fig:case_study}, we can see that SimANS indeed assigns larger probabilities to the negatives that rank near the positive while punishing the higher-ranking and lower-ranking ones that may be false negatives and uninformative negatives.
Furthermore, in \cref{case-b} where the positive is ranked at the first place, our approach is similar to the top-$k$ negatives sampling method that assigns larger probabilities to the higher-ranking hard negatives.

\section{Related Work}
Recent years have witnessed the remarkable performance of dense retrieval methods in text retrieval tasks~\cite{zhan2020repbert,hong2022sentence,ram2021learning,zhou2022debiased}.
Different from traditional sparse retrieval methods (\eg TF-IDF and BM25), dense retrieval approaches typically map queries and documents into low-dimensional dense vectors, and then utilize vector distance metrics (\eg cosine similarity) for retrieval. 

To learn an effective dense retrieval model, it is key to sample high-quality negatives paired with the given query and positives for training.
Early works~\cite{karpukhin2020dense,min2020ambigqa} mostly rely on in-batch random negatives and hard negatives sampled by BM25.
After that, a series of works~\cite{DBLP:conf/naacl/QuDLLRZDWW21,xiong2020approximate} find that sampling top-$k$ ranked examples by the dense retriever as hard negatives is more helpful to improve the retriever itself.
Among them, several methods~\cite{xiong2020approximate,DBLP:conf/sigir/ZhanM0G0M21} adopt a dynamic sampling strategy that actively samples top-$k$ hard negatives once after an interval during training.
However, these top-$k$ negative sampling strategies are easy to select higher-ranking false negatives for training.
To alleviate it, previous works have incorporated knowledge distillation~\cite{DBLP:conf/naacl/QuDLLRZDWW21,ren2021rocketqav2,lu2022ernie}, pre-training~\cite{zhou2022hyperlink,xu2022laprador} and other denoising techniques~\cite{mao2022curriculum,hofstatter2021efficiently}.
Despite the effectiveness, these methods mostly rely on complicated training strategies or complementary models.

In this work, we propose a simple but effective sampling method that weights the negative candidates with the consideration of their differences of relevance scores with positives. 
As a result, the ambiguous negatives with similar relevance scores to the positives will receive larger sampling probabilities, while the too hard (potential false negatives) and too easy negatives (uninformative) will be punished with smaller probabilities.

\end{document}